\documentclass[letterpaper, 10 pt, conference]{ieeeconf}  

\usepackage{balance}

\IEEEoverridecommandlockouts

\usepackage{amsmath, amssymb , graphicx, subcaption}
\usepackage{multirow}
\usepackage{dblfloatfix} 
\usepackage{soul}
\DeclareCaptionFont{caption}{\fontsize{8}{9.6}\selectfont}
\title{\LARGE \bf 
Augmented Reality Predictive Displays to Help Mitigate the Effects of Delayed Telesurgery}

\author{Florian Richter$^1$, Yifei Zhang$^2$, Yuheng Zhi$^3$, Ryan K. Orosco$^4$ \IEEEmembership{Member, IEEE}, \\and Michael C. Yip$^1$ \IEEEmembership{Member, IEEE}%
\thanks{$^1$Florian Richter and Michael C. Yip are with the Department of Electrical and Computer Engineering, University of California San Diego, La Jolla, CA 92093 USA. {\tt\small \{frichter, yip\}@ucsd.edu}}%
\thanks{$^2$Yifei Zhang is with the Department of Computer Science and Engineering, University of California San Diego, La Jolla, CA 92093 USA. {\tt\small yiz243@eng.ucsd.edu }}%
\thanks{$^3$Yuheng Zhi is with the Department of Computer Science and Engineering, Shanghai Jiao Tong University, Shanghai 200240, China {\tt\small zyh1996@sjtu.edu.cn}}%
\thanks{$^4$Ryan K. Orosco is with the Department of Surgery - Division of Head and Neck Surgery, University of California San Diego, La Jolla, CA 92093 USA. {\tt\small rorosco@ucsd.edu}}%
}

\begin{document}

\maketitle
\thispagestyle{empty}
\pagestyle{empty}

\begin{abstract}
Surgical robots offer the exciting potential for remote telesurgery, but advances are needed to make this technology efficient and accurate to ensure patient safety. Achieving these goals is hindered by the deleterious effects of latency between the remote operator and the bedside robot. Predictive displays have found success in overcoming these effects by giving the operator immediate visual feedback. However, previously developed predictive displays can not be directly applied to telesurgery due to the unique challenges in tracking the 3D geometry of the surgical environment. In this paper, we present the first predictive display for teleoperated surgical robots. The predicted display is stereoscopic, utilizes Augmented Reality (AR) to show the predicted motions alongside the complex tissue found in-situ within surgical environments, and overcomes the challenges in accurately tracking slave-tools in real-time. We call this a Stereoscopic AR Predictive Display (SARPD). To test the SARPD's performance, we conducted a user study with ten participants on the da Vinci\textregistered{} Surgical System. The results showed with statistical significance that using SARPD decreased time to complete task while having no effect on error rates when operating under delay.
\end{abstract}

\section{Introduction}

\begin{figure}[t]
	\centering
	\includegraphics[width=8.5cm]{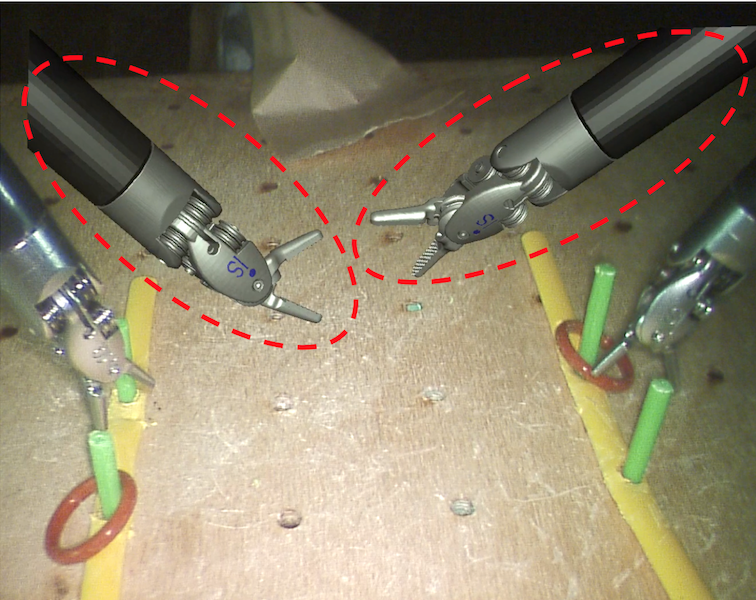}
	\caption{Stereoscopic AR Predictive Display (SARPD) showing the predicted slave-tools in real-time (highlighted with the dotted red lines), overlaid on the camera feedback under 1sec of round trip delay. SARPD can help users coordinate motor tasks preemptively to save time.}
    \label{fig:AR_Example}
\end{figure}

\begin{figure*}[b]
	\centering
	\includegraphics[width=16cm]{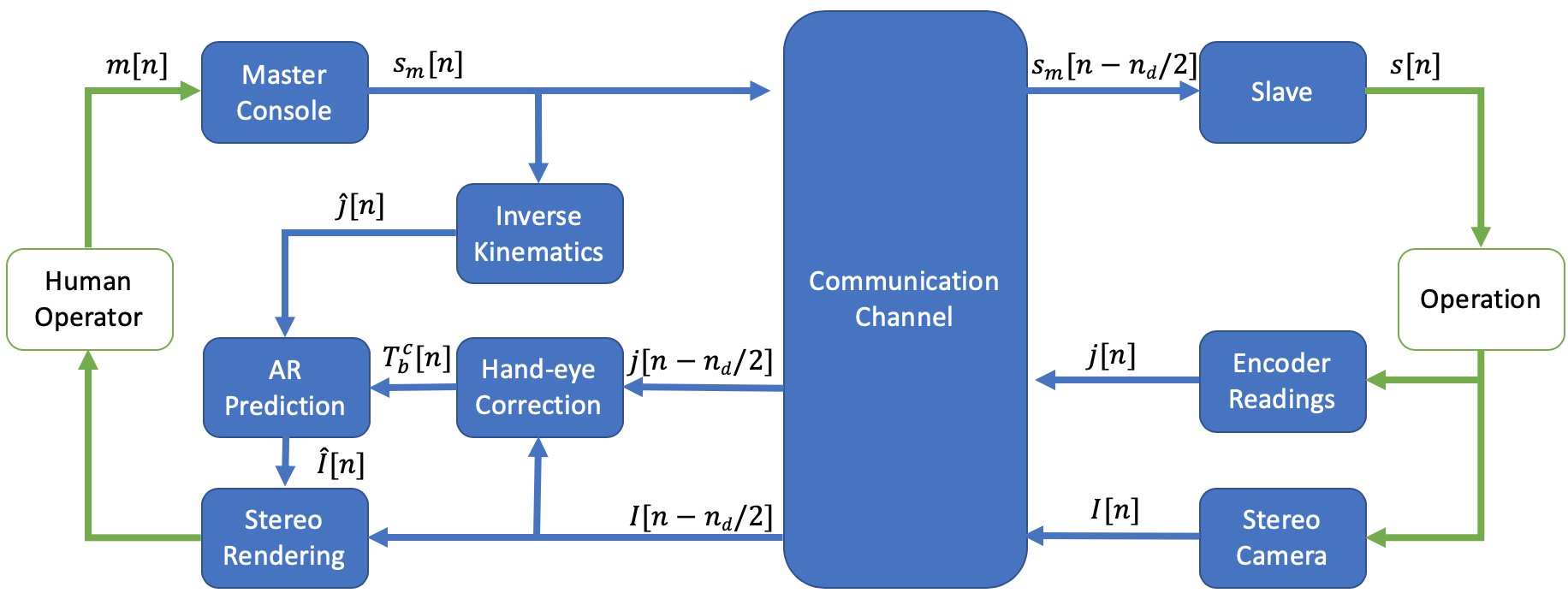}
	\caption{Architecture for SARPD with a round trip delay of $d$ where $m$, $s_m$, and $s$ are all poses, $j$ are the joint angles, $T^c_b$ are the hand-eye transforms between the base of the slave arm and camera, and $I$ are stereo images. For the sake of simplicity, the entire system is assumed to have a sampling rate of $f_s$, and let $n_d = f_sd$}
    \label{fig:FlowChart}
\end{figure*}

Teleoperation brings the advantage of remote control and manipulation to distant locations or harsh or constrained environments. The system allows operators to send commands from a remote console, traditionally called a $master$ device, to a robot, traditionally called a $slave$ device, and offers synchronization of movements. This allows the remote user to operate as if on-site, making teleoperational systems an ideal and often only solution to a wide range of applications such as underwater exploration, space robotics, mobile robots, and telesurgery \cite{yipDasJournal,TeleoperationHistory}.

Perhaps the most well known and still widely used teleoperation system to date is the da Vinci\textregistered{} Surgical System, which is deployed in thousands of hospitals. However, the master and slave systems are operated in the same room, providing some advantages of tremor reduction and stereoscopic vision \cite{reduceTremors} but none of the remote teleoperation functionality. The main technical challenge in realizing remote telesurgery (and similarly, all remote teleoperation) is the latency from the communication distance between the master and slave. Latency can reach beyond a second when linking sites between continents \cite{surgeryDelayMeasurement}, and may have higher ranges when using satellite or space communications. This delay causes overshoot and oscillations in the commanded positions, and are observable and statistically significant in as little as 50msec \cite{SteadyHand} \cite{TreatmentPlanning} of round trip communication delay. Anvari et al. reported on 21 remote laparoscopic surgeries where a distance of over 400km had delay of 135-140msec \cite{firstTeleRobotic}. Furthermore, 300msec has been stated as the largest delay where surgeons feel safe \cite{TransatlanticSurgery, LatencyEffects}, and when using satellite communication for operations between London and Toronto, a delay of 560msec was observed \cite{reasonForDelay}.

Teleoperational systems under delay is not specific to telesurgery and its history is too vast to be covered in this paper. Please refer to Hokayem and Spong's historical survey for a more extensive background on prior work \cite{TeleoperationHistory}. Nonetheless, early work by Ferrel and Sheridan suggested supervisory control to mitigate the negative effects of delay \cite{supervisoryControl}. This gives the robotic system the capability of making limited decisions on its own while being supervised by the operator; however, the master loses explicit control of the motions of the slave, and thus supervisory controlled can only be implemented practically at this time in structured environments. 

The largest area of research in delayed teleoperation has been for haptic feedback. When haptic feedback is given to the operator, it has been experimentally measured and theoretically shown that delay causes instability \cite{experimentalInstability, theoryInstability}. While there are some techniques to dampen the unstable overshoot and oscillatory behavior such as wave variables \cite{waveVariables}, these techniques have been shown to increase task completion time when under delay \cite{yipBadHaptic}, \cite{yipBadHaptic_2}, and teleoperating without haptic feedback can be often the better alternative. 

Predictive displays are virtual reality renderings, generally designed for space operations, that show a prediction of the events to follow in a short amount of time. It can be used to overcome the negative effects of delay by giving the operator immediate feedback from a predicted environment. Furthermore, it does not suffer stability issues that arise with delayed haptic feedback. Early predictive displays included manipulation of the Engineering Test Satellite 7 from ground control where the round trip delay can be up to 7sec \cite{spaceTeleopDelay2, spaceTeleopDelay3} and Augmented Reality (AR) rendering where the prediction is overlaid on raw image data \cite{AR_pd}. These strategies can be applied to telesurgery, but require overcoming the unique challenges in calculating and tracking the 3D environment for a full environment prediction, which includes non-rigid material such as tissue. Furthermore, prior work in the surgical robotics community highlights the need for active tracking rather than only relying on kinematic calibrations to localize the slave due to the millimeter scale of a surgical operation and the often utilized cable driven actuation \cite{need_tracking1, need_tracking2, EKF, need_tracking3, need_tracking4}.


In this paper we propose the first predictive display designed for teleoperated surgical robots. Most crucially, we calculate the predicted behavior of the robot arms and display this prediction in real-time to the operator to anticipate the delayed slave feedback video. The novel contributions of the paper are:
\begin{enumerate}
\item a real-time strategy for kinematically estimated AR registration, rendering, and lens-distorted image overlays for stereoscopic left and right streams of laparoscopic stereo cameras,
\item an Extended Kalman Filtering (EKF) strategy to address the challenges of visual-mismatch between the prediction and the actual movements that arise from imperfect kinematic calibrations with on-the-fly corrections, and
\item an adaptive transparency filter that prevents confusion arising from overlapping virtual and real visuals of the robot arms.
\end{enumerate} 

We present a complete system, and show through a user study that the result is an efficient AR rendering architecture for streaming stereoscopic displays. We call this a Stereoscopic AR Predictive Display (SARPD). We show that over 30fps for the stereoscopic AR rendering and 24Hz for the slave-tool tracking from the EKF can be achieved while running simultaneously on a commodity GPU and using the robot operating system (ROS). A user study is carried out to demonstrate its ability to improve the speed of procedures without affecting error rates. Beyond telesurgery, this solution can be deployed to any teleoperated robot with visual feedback once calibrated. Furthermore, it does not require stereo cameras or displays since the real-time slave-tool tracking is done on a single monocular camera data stream.  


\section{Methods}

A block diagram in Fig. \ref{fig:FlowChart} shows the architecture for SARPD, and the variables shown will be used throughout this paper. In surgical systems, such as the da Vinci\textregistered{} Surgical System, translational motions from the operator are scaled down to improve accuracy. This constant scaling relationship to set the slave's pose, $s$, from the operators input pose, $m$, in a teleoperation system under delay is described as follows:

\begin{equation}
	p_{s_m}[n] = scale(p_{m}[n] - p_{m}[n-1]) + p_{s}[n-1]
\end{equation}
\begin{equation}
	q_{s_m}[n] = q_{m}[n]
\end{equation}
\begin{equation}
	s[n] = s_m[n-n_d/2]
\end{equation}
where $p_x$ and $q_x$ are the translational component and quaternion of pose $x$. Equations (1) and (2) give the target pose for the slave, $s_m$, through the constant scaling for the position and mirroring the operators rotational input. Note that both $s_m$ and $s$ are poses in the corresponding slave arm base frame. The rotation is mirrored because master consoles such as the a Vinci\textregistered{} Surgical System use wrist orientation as an input. Equation (3) simply highlights the delay channel. 

The predictive display proposed here has two major components: slave-tool tracking by correcting the hand-eye transform in real-time to calculate an accurate prediction and stereoscopic AR rendering to display the prediction. Both are running asynchronously, and ROS is used to pass the data from the slave-tool tracking to the stereoscopic AR rendering.

\subsection{EKF to Correct Hand-Eye}

Ye et al. previously developed a tacking algorithm for the slave-tools and showed it to be accurate and robust to surgical environments \cite{EKF}. It successfully tracked the slave-tools in real-time by estimating the error of the initial hand-eye calibrations between the base of both slave arms and the left camera in real-time by using: 
\begin{enumerate}
	\item virtual slave-tool rendering to generate part-based templates online,
	\item template matching between virtual slave-tool parts and image data,
	\item geometric context to extract the best location estimates of the slave-tool parts, and
	\item EKF to track the error of the hand-eye transform using the previously found 2D estimates for the update.
\end{enumerate}

\begin{figure}[t]
    \vspace{2mm}
	\centering
	\includegraphics[width=8.5cm]{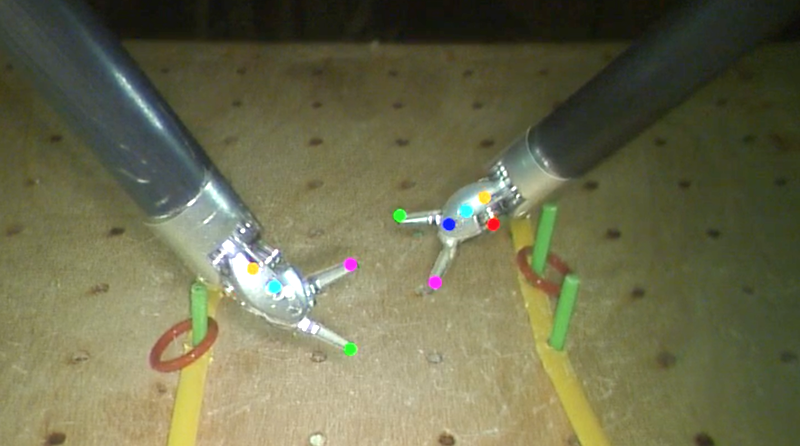}
	\caption{Implementation of hand-eye correction inspired by \cite{EKF} with adaptation for round trip delay of 1sec and parallelization to improve performance. Colored points represent visual features used in the EKF tracker that are related to tracking of kinematic robot links.}
    \label{fig:EKF}
\end{figure}

The hand-eye correction from \cite{EKF} is implemented on the master side, so the predicted AR slave-tools use the correction corresponding to the image data they will be displayed with. This means the joints angles, $j[n]$, from encoder readings and image data, $I[n]$, are passed through the communication channel before calculating the correction. This differs from \cite{EKF} which is not concerned with delayed teleoperation. Each slave-tool part is given its own thread in step 1 and 2 to improve the real-time performance. Fig. \ref{fig:EKF} is an example photo showing the detected features and rendered slave-tools that use the hand-eye correction. 

\begin{figure*}[t]
	\centering
	\vspace{2mm}
	\begin{subfigure}{.245\textwidth}
		\centering
  		\includegraphics[width=1\linewidth]{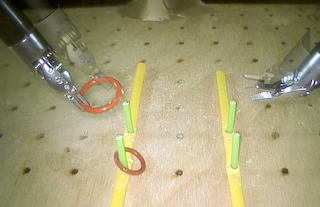}
	\end{subfigure}%
	\hspace{0.0001\textwidth}
	\begin{subfigure}{.245\textwidth}
  		\centering
  		\includegraphics[width=1\linewidth]{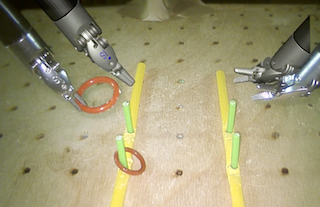}
	\end{subfigure}%
	\hspace{0.0001\textwidth}
	\begin{subfigure}{.245\textwidth}
		\centering
  		\includegraphics[width=1\linewidth]{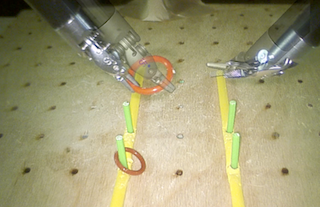}
	\end{subfigure}%
    \hspace{0.0001\textwidth}
   	\begin{subfigure}{.245\textwidth}
		\centering
  		\includegraphics[width=1\linewidth]{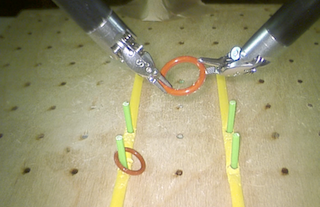}
	\end{subfigure}
	\caption{Example photos of AR prediction and dynamic transparency under 1sec of round trip delay. From left to right, a ring hand-off task is shown.}
    \label{fig:AR_Viewer_Fading}
\end{figure*}

To initialize the EKF, the hand-eye transform from calibration completes steps 1 through 3, and instead of step 4, a Perspective-n-Point (PnP) solver on the detected features is applied \cite{EKF}. To improve robustness, this is repeated when the distance or absolute difference in roll, pitch, or yaw between the corrected hand-eye and initial hand-eye from calibration is greater than a set threshold. This and the EKF updates to the corrected transform, $\hat{T}_b^c$, cause large steps in the hand-eye transform relative to the scale of surgical environments. These large steps will make the predicted AR slave-tool appear jumpy and therefore untrustworthy to operators. To smooth this, the following filter is applied to get the outputted hand-eye transform, $T_b^c$:

\begin{align}
	q_b^c[n] &= \frac{\text{sin}\big((1-a)\Omega[n]\big)}{\text{sin } \big(\Omega[n] \big)} q_b^c[n-1] + \frac{\text{sin}(a\Omega[n])}{\text{sin } \big(\Omega[n] \big)} \hat{q}_b^c[n]	\\
	\Omega[n] &= \text{cos}^{-1}\big(q_b^c[n-1] \cdot \hat{q}_b^c[n] \big)	\\
	p_b^c[n] &= (1-a) p_b^c[n-1] + a \hat{p}_b^c[n]
\end{align}
where $q_b^c$ and $p_b^c$ are the quaternion and translation representations of the outputted, smoothed hand-eye transform, $T_b^c$. Likewise, $\hat{q}_b^c$ and $\hat{p}_b^c$ are the corrected transforms from the EKF's output, $\hat{T}_b^{c}$. This is simply a first order infinite impulse response filter with parameter $a$ and spherical linear interpolation to average the rotations.

\subsection{Stereoscopic AR Rendering}

SARPD is rendered using The Visualization Toolkit (VTK) and OpenGL. To render the predicted slave-tools, their 3D CAD models are loaded as VTK Actors and uses the most recent filtered hand-eye transform to be in the left camera frame. The predicted joint angles for the slave-tools, $\hat{j}[n]$, are calculated through inverse kinematics from target pose $s_m[n]$, as shown in Fig. \ref{fig:FlowChart}. For the right display rendering, the slave-tools are additionally transformed using the baseline from stereo camera calibration to be in the right camera frame.

Distortion must be applied to the rendered slave-tools since the master console shows the distorted images to the operator. Camera calibration procedures in MATLAB and OpenCV find the coefficients for equations that un-distort barrel and tangential camera distortions. Solving for the inverse of these equations gives multiple solutions. Therefore, bilinear interpolation on the inverse of the discretized un-distort mapping is used to find the distortion map for both the left and right cameras.

Similar to how mappings are applied in OpenCV, the distortion map for each camera is split into two separate mappings, columns and rows. The four mappings are uploaded as textures onto the GPU once at the beginning and then applied to the slave-tool renderings using fragment shaders. 

Two virtual cameras, left and right, render the stereoscopic display. Each virtual camera display overlays the predicted and distorted slave-tool renderings on top of the corresponding image frame which are uploaded to the GPU as textures. Both the left and right rendering pipeline are done in parallel and the VTK slave-tool actors and textures are shared resources on the GPU. Through this implementation optimization and utilizing fragment shaders to apply the distortion mapping, the rendering pipeline can be run on consumer grade GPUs. Fig. \ref{fig:AR_Example} shows an example of the predicted slave-tools and image data rendering.


Even with the real-time hand-eye correction, there are other kinematic inaccuracies that are unaccounted for. Joint angle inaccuracies on the wrist of the slave-tool will cause inconsistency between the AR rendering and the image. This is noticeable when the slave-tools are making no motion. Even small, sub-millimeter inaccuracies were observed to cause confusion to users during initial studies. To overcome this, the opacity, $\alpha$, of the rendered slave-tools is dynamically set with the following equation:
\begin{align}
	l &= \big| \big| p_{s_m}[n] - p_{s} [n-n_d/2] \big| \big|										\\
	\alpha &= \text{min} \Big( \alpha_{max}, r\big(\text{max} ( l_{thresh},  l) - l_{thresh}\big) \Big)
\end{align}

The opacity will increase proportionally with the distance $l$, which is the distance between the new target pose that the prediction shows and the pose in the delayed image. The distance threshold, $l_{thresh}$, is the minimum value $l$ must have to not be fully transparent, and $\alpha_{max}$ simply sets the maximum opacity. The dynamic opacity allows for the operator to use the image data undisturbed when making precise, slow motions where the kinematic inaccuracies would be apparent and naturally use the AR prediction when making larger motions. Fig. \ref{fig:AR_Viewer_Fading} shows an example of the dynamic transparency during a ring hand-off.

\section{Experimental Setup}

To initialize the hand-eye transform for the EKF, we provide a calibration method that only requires a rigidly mounted checkerboard on the slave-tools (only during the calibration phase). This is the only additional hardware required to implement the proposed predictive display, and calibration is only required once before a procedure. Performance and latency tests were also conducted to show the efficiency of SARPD.

To measure the effectiveness SARPD, a user study with ten participants was conducted on the da Vinci\textregistered{} Surgical System. The participants were seven novices and three surgeons who use the da Vinci for their practice. Errors and time to complete task are used as metrics, and statistical testing was done to draw conclusions of SARPD's performance. A recent study found that when using satellite communication for telesurgery between London and Toronto, a round trip delay of $560.7 \pm 16.5$mecs was measured \cite{reasonForDelay}. So a round trip delay of 1sec was used for the delayed environment in the user study to ensure SARPD performs in realistically high latency remote operations.  

SARPD is rendered to the master console at 30fps. The operator manipulates two master arms in the console to get pose $m$. There are two 7-DOF slave arms which go to pose $s$. Modifications were made to the daVinci Research Toolkit (dVRK \cite{dVRK}) to support the architecture proposed in Fig. 2 for the da Vinci\textregistered{} Surgical System. The computer used to run the modified dVRK and SARPD has an Intel\textregistered{} Core\texttrademark{} i9-7940X Processor and NVIDIA's GeForce GTX 1060.


\begin{figure*}[b]
	\centering
	\begin{subfigure}{.24\textwidth}
  		\centering
  		\includegraphics[width=1\linewidth]{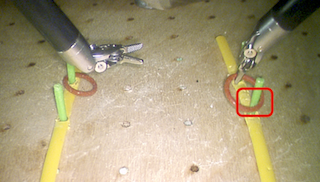}
	\end{subfigure}
	\begin{subfigure}{.24\textwidth}
		\centering
  		\includegraphics[width=1\linewidth]{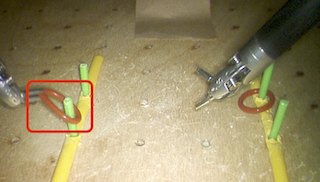}
	\end{subfigure}
	\begin{subfigure}{.24\textwidth}
  		\centering
  		\includegraphics[width=1\linewidth]{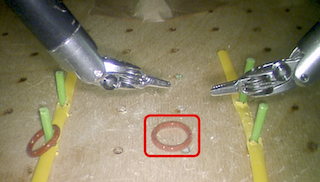}
	\end{subfigure}
	\begin{subfigure}{.24\textwidth}
		\centering
  		\includegraphics[width=1\linewidth]{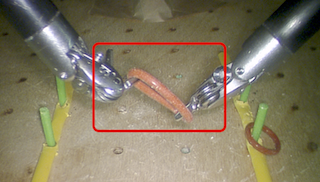}
	\end{subfigure}
	\caption{Example errors participants made. From left to right: touch peg, stretch ring on peg, drop ring, and stretch ring during handoff.}
    \label{fig:exampleErrors}
\end{figure*}

\subsection{Initial Hand-Eye Calibration}

To find the initial hand-eye transform, a checkerboard is rigidly attached to the end-effector of the slave-tool. The arm is moved around and photos are taken of the checkerboard and the corresponding joint angles are recorded. The $i$-th corner on the checkerboard in the checkerboard frame, $p_e(i)$, with side length $s$, is projected onto the image plane with the following equation: 

\begin{equation}
	 p_c (i, j) = \frac{1}{d} K T_b^c T_{-e}^b (j) T_e^{-e} p_e(i)
\end{equation}

$\frac{1}{d}K$ projects a point onto the image plane using the camera matrix. $T_{-e}^b(j)$ is the pose of the checkerboard which is calculated by using forward kinematics from the recorded joint angles at image $j$, and $T_e^{-e}$ is the constant offset error of the calculated $T_{-e}^b (j)$ which accounts for mounting errors of the checkerboard. To solve for the hand-eye transform in a robust manner, the following optimization problem is solved when there are $n$ corners on the checkerboard and $m$ recorded images and joint angles:

\begin{equation}
	\text{ }\underset{T_b^c, T_e^{-e},s}{\operatorname{argmin}}\text{ } \frac{1}{m} \sum \limits_{j=1}^m \sum \limits_{i=1}^n \Big|\Big| p_c(i,j) - z(i,j) \Big|\Big|^2
\end{equation}
where $z(i,j)$ is the pixel position of the checkerboard corner on the recorded image. Optimizing the constant offset of the checkerboard and side length, $s$, will help account for alignment errors and printing inaccuracies. $T_e^{-e}$ is initially set to the identity matrix.

\begin{figure}[t]
    \vspace{2mm}
	\centering
	\includegraphics[width=8.5cm]{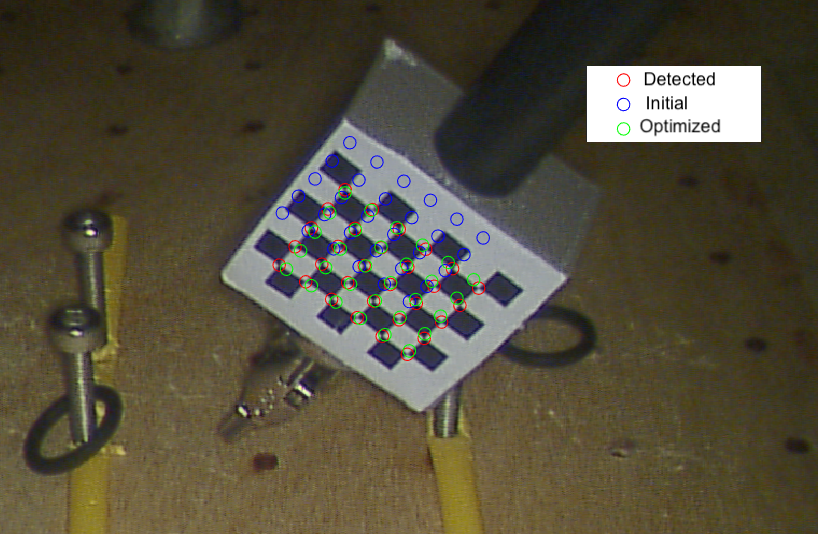}
	\caption{Calibration tool on the instruments to find the transform from the slave arms base to the camera frame. This calibration is carried out once, as needed to attain an initial hand-eye transform for the EKF. The circles highlight the re-projected checkerboard corners.}
    \label{fig:hand-eye}
\end{figure}

To initialize the hand-eye transform, a PnP solver is used on each recorded image to find the pose of the camera in the checkerboard frame. This is then further transformed with the corresponding $T_{-e}^b(j)$ to get image $j$'s individual estimate of the hand-eye transform. The initial value for the hand-eye is simply set to the arithmetic average of the positional component and roll, pitch, and yaw of all of the images individual estimates. The optimization is solved by using MATLAB's fmincon function, and Fig \ref{fig:hand-eye}. shows an example result of the re-projected corners on a recorded image.

\subsection{Task for User Study}

\begin{figure}[t]
    \vspace{2mm}
	\centering
	\includegraphics[width=8.5cm]{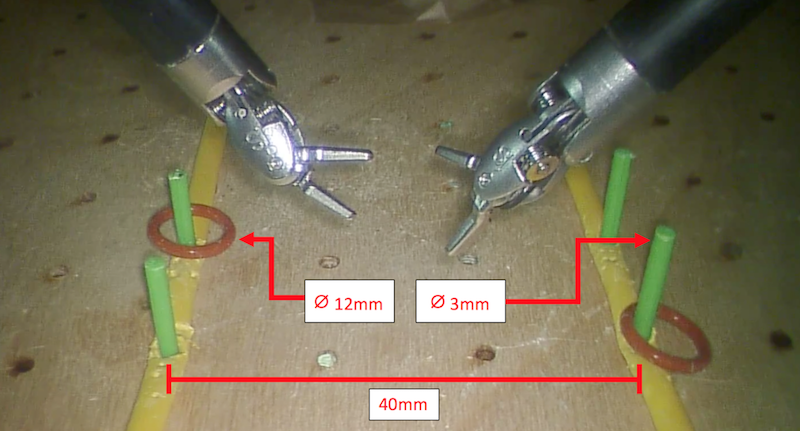}
	\caption{Environment setup used for the peg transfer task. Transfer of rings are carried out between instruments and between rows of pegs to characterize single and dual arm control in pickup, hand-off, and place-down task.}
    \label{fig:task_environment}
\end{figure}

A peg transfer task, similar to a task in the Fundamentals of Laparoscopic Surgery, is the sample task used in the user study due to the complex motions involved with it. A photo of the environment from the endoscope is provided in Fig. \ref{fig:task_environment}. To complete the task, the operator must pick up the ring from the front right or left peg with the corresponding arm, pass the ring to the other arm, place the ring on to opposite front peg, and then repeat with the back pegs and ring.

Time to complete task and weighted error are used to evaluate performance of completing the task. To find the weighted error, all errors that occurred during a trial are counted and weighted according to Table 1. Example errors are shown in Fig. \ref{fig:exampleErrors}. The weightings were chosen such that the severity of the errors would be reflected properly when evaluating the performance of completing the peg transfer. Each participant was also shown a video of a complete task with no errors and examples of the errors before starting. 

\begin{table}[h]
\caption{\\Weights associated with type of error}
\begin{center}
\begin{tabular}{ |c|c| } 
\hline  
	\textbf{Error} & \textbf{Weight}								\\   \hline
	Touch peg & 1				 							\\   \hline
	Touch ground & 2					 					\\   \hline
	Stretch ring during handoff for a second or less & 2 			\\   \hline
	Drop ring & 3					 						\\   \hline
	Stretch ring on peg for a second or less & 4					\\   \hline
	Stretch ring for an additional second & 4					 	\\   \hline
	Stretch/move peg & 10					 				\\   \hline
	Knock down peg & 20					 				\\   \hline
\end{tabular}
\end{center}
\end{table}

\begin{figure*}[b]
	\centering
	\begin{subfigure}{.48\textwidth}
  		\centering
  		\includegraphics[width=1\linewidth]{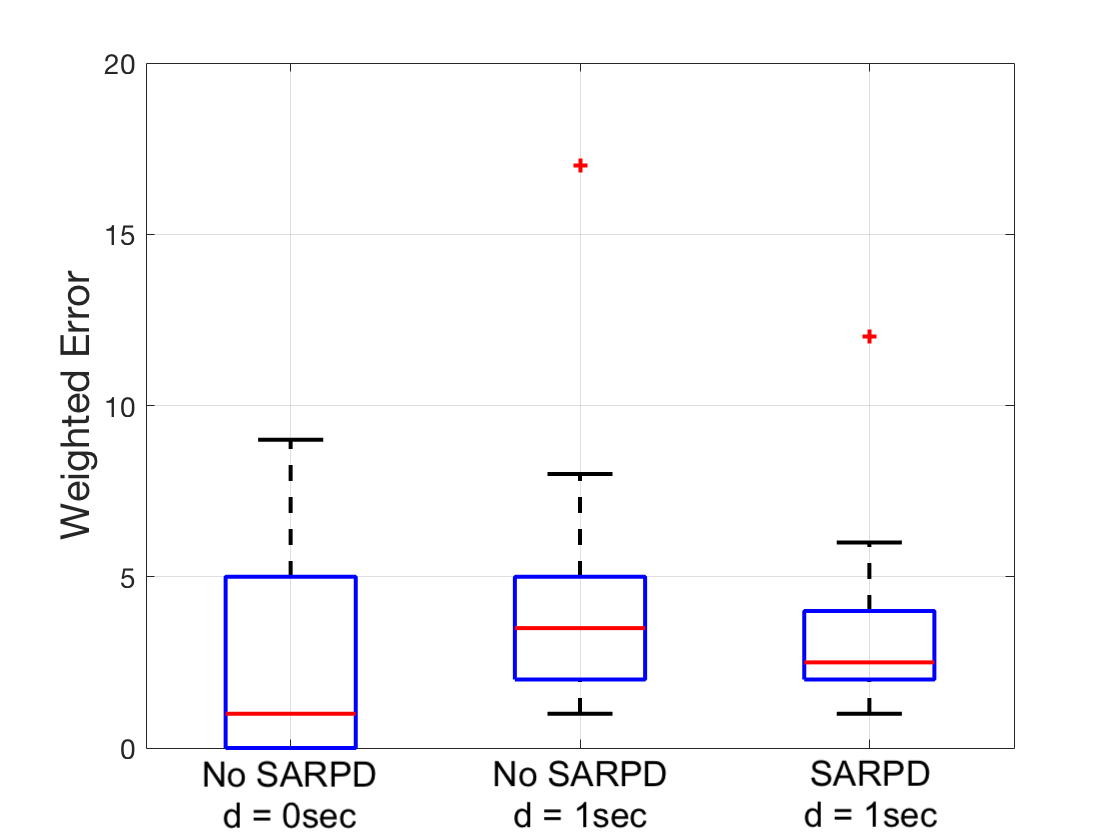}
  	\caption{Weighted Error}
	\end{subfigure}
	\begin{subfigure}{.48\textwidth}
		\centering
  		\includegraphics[width=1\linewidth]{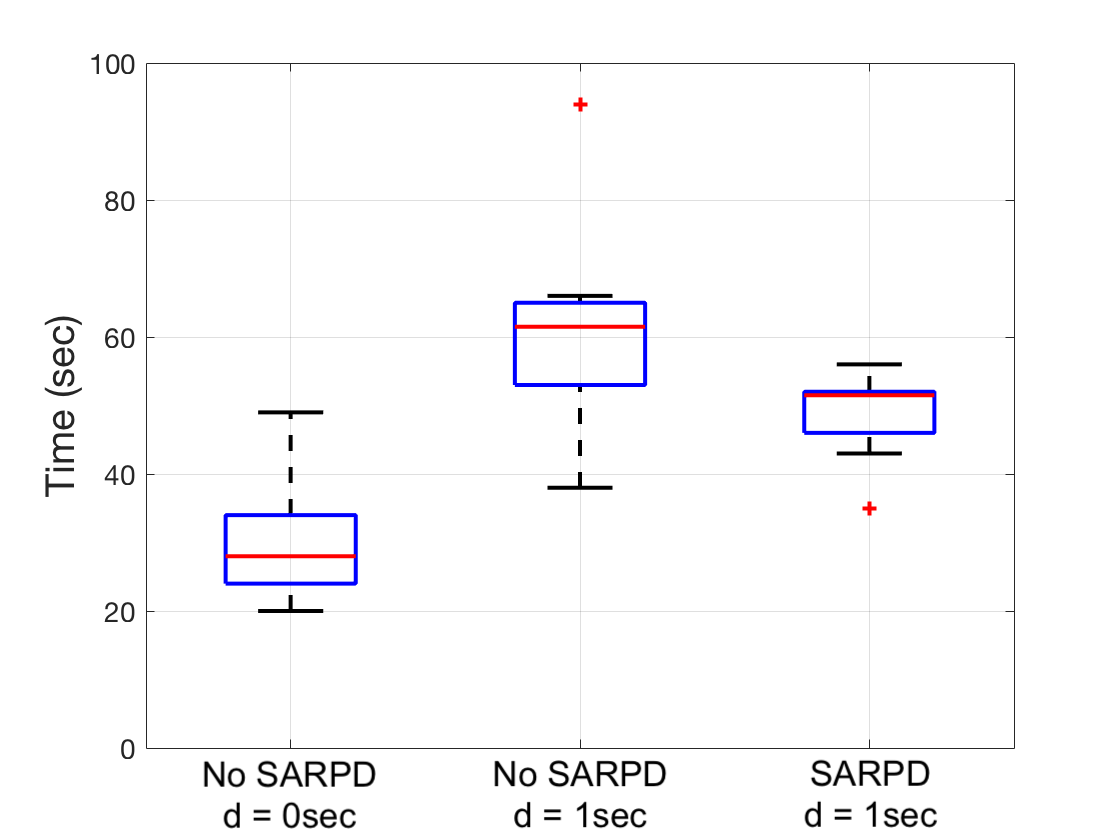}
  		\caption{Time to Complete Task}
	\end{subfigure}
	\caption{Weighted error and completion time results from the user study. No statistically significant change in weighted errors is observed with SARPD, but a statistical decrease in completion time is observed when teleoperating under delay.}
    \label{fig:userStudyPlots}
\end{figure*}

\subsection{Procedure for User Study}

After introducing the participant to the study and showing the video, the participant goes through the following procedure:

\begin{enumerate}
	\item Practice: complete the task twice under no delay and with no SARPD
	\item Record: complete the task once under no delay and with no SARPD
	\item Repeat step 1 and 2 under delay and with no SARPD
	\item Repeat step 1 and 2 under delay and with SARPD
\end{enumerate}

Two practice trials are always completed before every recording, so that the participant can overcome the learning curve for the new environment. To further ensure this, participants are also offered additional practice before recording. The order was chosen so that participants only encounter one new environmental effect at a time. Through initial experimentation we set the following values: $scale = 0.2$ (the default value for dVRK) for equation (1), $a = 0.8$ for equation (4) and (6), and $l_{thresh} = 5.3$mm, $\alpha_{max} = 0.8$, and $r = 100$ for equation (8).

Note that SARPD without using the active tracking from the EKF is not included in the experiment. This was elected due to the the highly inaccurate prediction from the AR rendering when relying on the kinematic calibration. 

\section{Results}

The original work for the slave-tool tracking \cite{EKF} was measured to run at 26Hz and 13Hz for the part-based template generation and overall tracking respectively. Through the parallelization efforts, the slave-tool tracking now runs at 50Hz and 24Hz for the part-based template generation and overall tracking respectively. During this measurement, the stereoscopic AR rendering ran asynchronously at 36fps.



We also measured the latency of the entire image pipeline, which is the time from when an image is captured to when it is displayed to the operator under no delay. When using the SARPD, the latency was measured to be $100 \pm 20$ms. To compare, we ran the same experiment with ROS's $image\_view$ to display the raw image data and measured a latency of $110 \pm 20$ms. ROS's $image\_view$ is the recommended way to view images in dVRK, so our solution also improves on the latency of the system. 


\begin{table}[h]
\centering
\caption{\\Average and standard deviation from the user study results}
\begin{tabular}{|c|c|c|}
	\hline
							&  Weighted Error 		& Completion Time (sec) 			\\ \hline
	 \textbf{No SARPD, d=0sec}		& $2.8\pm3.4$ 	& $29.6\pm8.8$	\\ \hline
	 \textbf{No SARPD, d=1sec}		& $4.9\pm4.7$ 	& $60.6\pm14.8$	\\ \hline
	 \textbf{SARPD, d=1sec}		& $3.7\pm3.2$ 	& $49.2\pm6.3$	\\ \hline
	\end{tabular}
\end{table}

The results of the user study comparing delayed teleoperation with and without SARPD is shown in Fig. \ref{fig:userStudyPlots} and the statistics are given in Table 3. To identify a statistically significant ($p<0.05$) change in weighted error or time to complete task under the different conditions, two sided paired t-test's were done. No statistically significant difference was measured in weighted errors when comparing SARPD under delay with not using SARPD under both no delay ($p=0.55$) and delay ($p=0.50$). Repeating for time to complete task, statistical significance was measured when comparing SARPD under delay with not using SARPD under both no delay ($p=1.7e-4$) and delay ($p=0.02$). To conclude, we measured that using SARPD on average decreases the time to complete task by 19\% when under delay while having no statistically significant change in errors.

\section{Discussions and Conclusion}

SARPD is the first predictive display developed for teleoperated surgical robots and has been shown to be an efficient system. On a consumer grade GPU, the tool-tracking is computed in real-time and the AR rendering pipeline is able to support two 1080p displays, each running above 30fps. By utilizing AR to show the prediction, no assumptions need to be made about the environment so SARPD can be applied to any calibrated teleoperated system under delay. SAPRD also on average decreases time to complete task while not affecting the number of errors. This is an expected result since the AR rendering should not assist when small, precise motions are made which is when the errors in our user study occur.

For future work, more studies need to be done on different surgical tasks with rigorous accuracy measurements and subjective data to understand cognitive load when operating with and without SARPD under various levels of delay. In particular, tasks with more environmental interaction. Furthermore, additions to the tracking algorithm must be done to incorporate endoscopic motions which is part of normal operating procedure. The rendering pipeline also needs to account for obstructions of the the slave-tools from environmental obstacles. 


\section{Acknowledgements}
We thank all the participants from the user study for their time, all the members of the Advanced Robotics and Controls Lab at University of California San Diego for the intellectual discussions and technical help, and the UCSD Galvanizing Engineering in Medicine (GEM) program and the US Army AMEDD Advanced Medical Technology Initiative (AAMTI) for the funding.\\
\balance
\bibliographystyle{ieeetr}
\bibliography{references}

\end{document}